\documentclass{article}

\usepackage{arxiv}

\usepackage[utf8]{inputenc} 
\usepackage[T1]{fontenc}    
\usepackage{hyperref}       
\usepackage{url}            
\usepackage{booktabs}       
\usepackage{amsfonts}       
\usepackage{nicefrac}       
\usepackage{microtype}      
\usepackage{lipsum}
\usepackage{graphicx}
\graphicspath{ {./images/} }

\title{Intelligent methods for business rule processing: State-of-the-art}

\author{
 Cristiano André da Costa, Uélison Jean Lopes dos Santos, Eduardo Souza dos Reis,\\ \textbf{Rodolfo Stoffel Antunes, Henrique Chaves Pacheco, Thaynã da Silva França} \\
 \textbf{Rodrigo da Rosa Righi, Jorge Luis Victória Barbosa} \\
  Software Innovation Laboratory - SOFTWARELAB\\
  Universidade do Vale do Rio dos Sinos - UNISINOS\\
  \texttt{\{cac,uelisonj,esreis, rsantunes, hcpacheco, tfrancanan, rrrighi, jbarbosa\}@unisinos.br} \\
   \And
 Franklin Jebadoss, Jorge Montalvao, Rogerio   Kunkel \\
  Dell Digital Services\\
  Dell EMC\\
  \texttt{\{franklin\_jebadoss, j\_Montalvao, Rogerio\_Kunkel\}@dell.com} \\
}

\begin{document}
\maketitle
\begin{abstract}
In this article, we provide an overview of the latest intelligent techniques used for processing business rules. We have conducted a comprehensive survey of the relevant literature on robot process automation, with a specific focus on machine learning and other intelligent approaches. Additionally, we have examined the top vendors in the market and their leading solutions to tackle this issue.
\end{abstract}


\section{Introduction}
Business automation processes have gained popularity in recent times. Robot Process Automation (RPA) reached its peak in September 2018, according to Google Trends data~\cite{syed2020robotic}. In this article, we provide an in-depth analysis of selected papers that describe the current state-of-the-art on RPA and Intelligent Process Automation (IPA).

The main objective of this article is to present the latest research and understanding of intelligent methods for processing business rules, especially related to service order handling. The methods discussed involve the use of machine processing techniques and natural language processing.

The article is structured as follows: Section 2 describe the research methodology. Section 3 focuses on Robot Process Automation (RPA). Section 4 discusses Intelligent Process Automation (IPA). Section 5 explains the machine learning approaches to IPA. Section 6 presents the leading vendors of RPA and IPA solutions. Finally, in Section 7, we draw conclusions based on our research.

\section{Method}
\label{sec:method}
To begin our research process, we needed to select specific keywords and define the scope of our search. To accomplish this, we utilized keywords associated with RPA, IPA, and CPA and searched through Google Scholar~\footnote{https://scholar.google.com} and Scopus \footnote{https://www.scopus.com} to identify relevant articles. After conducting additional filtering and exclusion processes, we removed works that did not fully address our research questions from the corpus. The initial search yielded 107 articles, but after filtering, we thoroughly reviewed 34 articles.

In the following sections, we will outline the most significant findings from recent studies on process automation, particularly RPAs, and discuss the latest advancements in IPAs, which incorporate cutting-edge techniques such as machine learning, deep learning, and generative modeling.

\section{Robot Process Automation}
\label{sec:rpa}

Currently, the literature refers to RPAs as the set of tools used to develop bots, which mimic human behavior on business tasks, typically process-aware tasks~\cite{van2018robotic,syed2020robotic}.
For instance, opening a spreadsheet, changing values, and committing.
Bots bring many advantages to human agents, among which we highlight enabling humans to make only creative, social, and decision-making tasks~\cite{van2018robotic};
accuracy can be expected to reach 100\%, and availability of 24/7 working schedule~\cite{syed2020robotic}; cost reduction of the process (a standard RPA costs a third of the cost of a full-time employee~\cite{aguirre2017automation}); RPAs are highly scalable to meet a varying intensity of demands~\cite{wewerka2020robotic}; and bots allow for transparent and detailed documentation, increasing compliance~\cite{wewerka2020robotic}.

Another essential characteristic of an RPA is its non-intrusiveness, meaning its adoption and integration with business tools should be as effortless as possible~\cite{hindel2020robotic}.
In turn, social skills are more suited for human agents, demanding empathy and social interactions in order to build trust and customer relationships~\cite{wewerka2020robotic}.
In other words, the critical idea of RPA is to allow the workers to spend more time on decision-making tasks instead of repetitive manual labor, which can be replaced by a bot~\cite{aguirre2017automation}.
Formally, the IRPA-AI Institute~\footnote{https://irpaai.com/ available on April 2023} definition of RPA is: 

\begin{enumerate}
    \item[$\bullet$] \textit{``the application of technology allowing employees in a company to configure computer software or a `robot' to capture and interpret existing applications for processing a transaction, manipulating data, triggering responses, and communicating with other digital systems''}.
\end{enumerate}

Quantitatively, Syed et al\cite{syed2020robotic} report that RPA technology has proven to cut the cost of human resource-related spending
by 20–50\% and a significant reduction (from 30\% to 70\%) in the process cycle time.
\cite{aguirre2017automation} experiments show an improvement of 21\% on the number of cases agents can handle while supported by an RPA.
Consequently, in the last five years, RPA became mainstream, consisting of a set of tools that can automate processes based on business rules, being classified as a highly promising approach~\cite{wewerka2020robotic,aguirre2017automation}.

Within a business RPA tool, there are many modules and multiple bots. Yet, an overview was defined by \cite{syed2020robotic}, consisting of three main components: a graphical modeling tool, an orchestrator, and the bot itself. 
Graphical modeling tools allow configuration and customization of the bots' features by the human agent.
In a complex process, more than a single bot (defined per task) is needed, and a collection of dependent bots must be developed, thus the need for an orchestrator.
The orchestrator is the software responsible for the monitoring and availability of the system as a whole, composed of a task scheduler and an urgency metric, to enable prioritization of resources to a given bot when the system is overloaded.
Other standard components are performance analytic tools, which enable the evaluation and comparison of the bots' work with human agents to contrast their efficiency. 

Aguirre and Rodriguez \cite{aguirre2017automation} propose criteria to identify which tasks to automate, choosing highly structured tasks, usually back office.
Other proposed criteria are:

\begin{itemize}
\setlength\itemsep{1em}
  \item Low cognitive requirements. A task that does not require subjective judgment, creativity, or interpretation skills.
  \item High frequency. A task that is repeated constantly during the human agent's work cycle.
  \item Access to multiple systems. A process that requires access to multiple applications and systems to perform the job.
  \item Limited exception handling. Tasks that have limited or no exceptions to handle.
  \item Prone to human errors.
\end{itemize}

One of the considered works \cite{rizun2020discovery} includes IT ticket automation as a core task for RPAs in the enterprise context.
They are usually composed of unstructured (client interaction logs) and structured (agent's notes) text.
Therefore, natural language processing is essential in IT ticket processing, offering a broad spectrum of project opportunities.

\begin{figure}[t]
    \centering
    \includegraphics[width=0.8\linewidth]{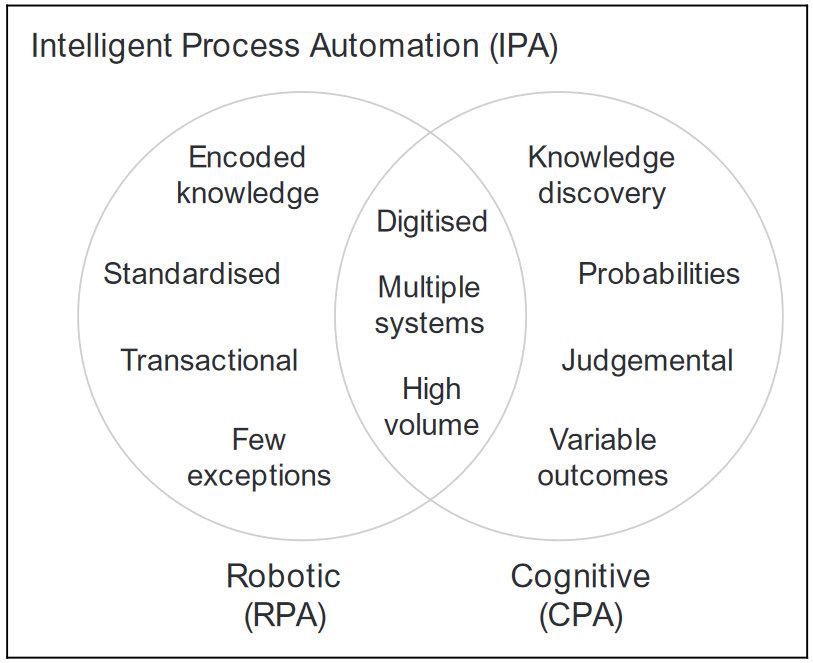}
    \caption{The IPA area is defined in the literature as the intersection between RPA and Cognitive Process Automation~\cite{richardson2020cognitive}. Consulting firms like Deloitte and Capgemini argue that the main areas where RPA can be applied are accounts payable, accounts receivable, travel and expenses, fixed assets, and human resource administration. With cognitive strategies, it is possible to address intricate problems that deal with probabilities and pattern recognition~\label{fig:ipa_diagram}.}
\end{figure}

\section{Intelligent Process Automation}
\label{sec:ipa}

The tasks performed by RPAs perform are typically rule-based, well-structured, and repetitive~\cite{syed2020robotic}.
Nevertheless, future RPAs or IPA, must include modules capable of dealing with unstructured data~\cite{van2018robotic}.
In contrast, IPAs are more expensive to build.
One prevalent application of IPAs is the natural language processing bots that, combined with machine learning, will replace human agents in customer relations activities~\cite{syed2020robotic}.

An IPA adds another layer to the standard RPA: Cognitive Process Automation (CPA), as depicted in Figure~\ref{fig:ipa_diagram}.
CPA focuses on knowledge work and utilizes constructed AI instead of classical AI.
Moreover, CPA can integrate with deep learning to incorporate natural language generation, computer vision (AI-screen recognition), and self-improvement~\cite {richardson2020cognitive}.
Improving over time and changing the decision-making process as the actual business process changes are the main feature of current IPAs, yet remains largely an open issue.
Online learning is achieved through monitoring and retraining the models~\cite{chakraborti2020robotic}.
The objective is to enable IPAs to identify such changes, predict future risks, and either alert or adapt when feasible.
In this context, generative models are suited for validating, monitoring, and adapting models to novel situations.

\section{Machine Learning approaches to IPA}

When it comes to machine learning, IPAs (Intelligent Personal Assistants) utilize online learning, which involves continuous changes in model weights, to process complex inputs. These inputs typically consist of unconstrained textual information mixed with categorical elements annotated by a human agent, such as urgency, observations, or required parts.

Most tasks can be accomplished using a sequence-to-sequence modeling framework, which is especially useful for natural language processing. This framework involves the mapping of input tokens, such as words, to output tokens by way of similarity. Essentially, this means mapping the probability distribution of the next token in the output sequence to a variable-length input sequence.

In practice, the mapping is not done directly, and the model also yields intermediate representations that encode the context in which the input tokens are presented. These representations are referred to as context vectors in natural language processing, and the two-step procedure is known as encoder-decoder methods~\cite{gru2014}.

Therefore, the set of input tokens is a vector of words encoded in a continuous representation, such as word2vec~\cite{word2vec2013}.

\begin{figure}
    \centering
    \includegraphics[width=0.85\linewidth]{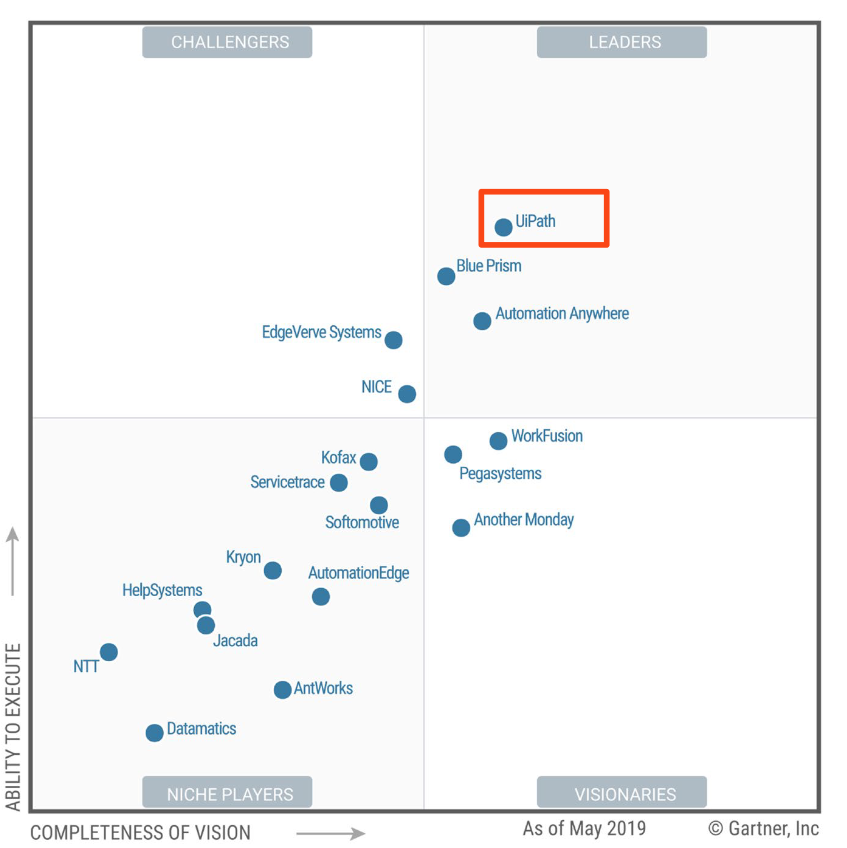}
    \caption{Gartner's magic quadrant for RPA shows UiPath, Automation Anywhere and Blueprism as the leaders in the sector~\cite{syed2020robotic}.}
    \label{fig:magic_quadrant}
\end{figure}

Initially, the community leaned towards recurrent models for both decoders and encoders, mainly to the RNNs~\cite{universalTransformers2018}.
Nonetheless, RNN-based models have visible shortcomings.
For one, relying on a unidirectional pipeline, the network can access the full context at the end of a sentence. Still, earlier iterations need to have information on the incoming tokens.
Instead of increasing the length of the context vector, the Transformers~\cite{transformers2017} selectively look for the most informative tokens at each timestep through self-attention.
Self-attention is an attention mechanism that correlates different positions of the same sequence, given higher weights for correlations among tokens within the same context, due to referring to the same object or being close together in the sentence.
Through self-attention, each cell of the context vector is informed by all previous inputs, resulting in a sizeable receptive field over the whole sentence~\cite{universalTransformers2018}.

In contrast, for categorical objects, classifiers such as Neural Networks or Random Forests have been widely explored. 
Therefore, an ideal IPA must implement different machine learning models to act over other domains on each bot.
As the number of models increases, the communication among them becomes a bottleneck. Thus, models capable of projecting the heterogeneous inputs in a single vectorized encoding are needed.
In other words, the goal is to use the heterogeneous input projection to a regularized subspace of simple N-dimensional vectors as a preprocessing step and then feed this vector representation to the actual classifiers or decision models. 

\section{Vendors}
\label{sec:vendors}

In recent years, RPA has emerged as the solution for process automation. The RPA market is predicted to reach a market volume of \$ 2.9 billion in 2021~\cite{hindel2020robotic}. As a result, a large number of vendors proposed their tools, creating a challenging competition between players. Market leaders are Blue Prism, arguably the pioneer RPA product, UIPath, and Automation Anywhere~\cite{syed2020robotic}.
All three leaders include modern techniques, such as OCR, for extracting data from documents, and Computer Vision, allowing the bots to interact with objects on HTML, PDF, or virtual desktop interfaces. 
Another key component is understanding the unstructured text consumers use when interacting with the bots.
Therefore, natural language processing became central in the RPA market, automating chatbots, form filling, and voice interactions.
Other widely mentioned products on our corpus include Workfusion, Kryon Systems, Softomotive, Contextor, EdgeVerve, niCE, and Redwood Software.
Pegasystems and Cognizant provide RPA functionality embedded into more traditional CRM and BI functionalities, providing less intrusiveness.
Figure~\ref{fig:magic_quadrant} classifies most of the studied products by distinct features.

When RPA exploded in popularity, some important issues, such as scalability, interoperability, and portability, emerged. Since a large portion of enterprise applications are developed based on graphical user interfaces (GUI), some vendors are developing specific APIs to facilitate portability between client software and bots. Furthermore, according to Simek et al.\cite{vsimek2019robot}, multiple RPA solutions can be integrated via these APIs. 

\begin{figure}
    \centering
    \includegraphics[width=0.9\linewidth]{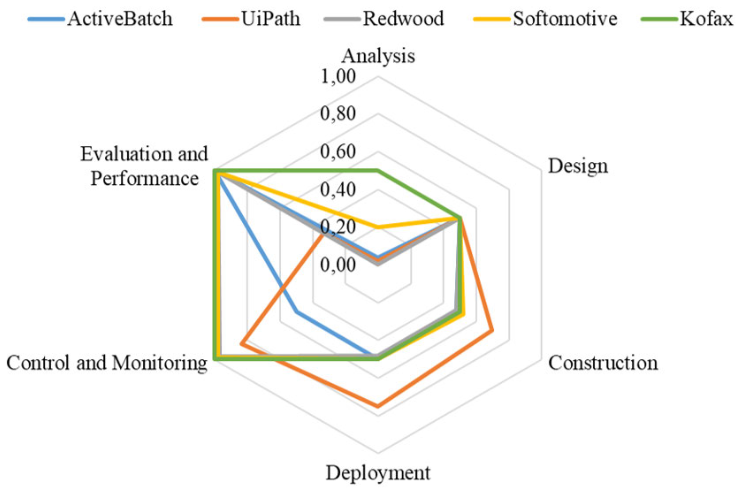}
    \caption{Comparison by development steps of the most complete vendor tools~\cite{enriquez2020robotic}.}
    \label{fig:vendor_comparison_diagram}
\end{figure}

As for performance metrics, most businesses rely on quality metrics that are manageable to generalize.
Hence, valuable indicators such as resource allocation and execution time are more straightforward metrics for IPAs~\cite{richardson2020cognitive}.
Other options for the current project context include benchmarks, such as management ticket texts provided in~\cite{rizun2020discovery}.
Based on the same RPA development methodology that we adopted, Figure~\ref{fig:vendor_comparison_diagram} compares the leading players by efficiency on each step.

\section{Conclusion}
\label{sec:conclusion}
This article provides a literature review of RPA and its related concepts such as Cognitive Process Automation and IPA. The review helps to understand the current issues and possible technological applications of RPA from both scientific and business perspectives. We also compare the existing RPA vendors.

The existing literature on RPA primarily focuses on definitions and case studies, while studies on technical and implementation strategies are scarce. To address this gap, future research could combine machine learning and pattern recognition with RPA, which some researchers refer to as IPA. It is crucial for RPA strategies to integrate with current systems, other tools, and technologies to provide scalability and performance according to business needs. Therefore, future studies must combine these technologies.

\bibliographystyle{unsrt}

\end{document}